\documentclass{article}
\usepackage[utf8]{inputenc}
\usepackage{comment}
\usepackage{todonotes}
\usepackage{url}

\title{Untangling the GDPR Using ConRelMiner}

\author{Karolin Winter, Stefanie Rinderle-Ma\\
Faculty of Computer Science, University of Vienna, Vienna, Austria \\
\{karolin.winter, stefanie.rinderle-ma\}@univie.ac.at}
\date{}
\begin{document}

\maketitle

\begin{abstract}
The General Data Protection Regulation (GDPR) poses enormous challenges on companies and organizations with respect to understanding, implementing, and maintaining the contained constraints. We report on how the ConRelMiner method can be used for untangling the GDPR. For this, the GDPR is filtered and grouped along the roles mentioned by the GDPR and the reduction of sentences to be read by analysts is shown. Moreover, the output of the ConRelMiner -- a cluster graph with relations between the sentences -- is displayed and interpreted. Overall the goal is to illustrate how the effort for implementing the GDPR can be reduced and a structured and meaningful representation of the relevant GDPR sentences can be found. 
\end{abstract}

\section{Introduction and the ConRelMiner Method}
Providing support for analyzing regulatory documents is of utmost importance for many companies nowadays as they face constantly changing or new requirements such as recently the General Data Protection Regulation (GDPR)\footnote{\url{https://eugdpr.org/}}. Nowadays this is mostly done in a manual way which can be error-prone and costly. Hence, our recent research (cf.\cite{WiRi2018,DBLP:conf/otm/WinterRGFM17}) aims at providing (semi-)automatic means to analyze regulatory documents based on text and data mining methods. 

\begin{figure}[htb!]
    \centering
    \includegraphics[scale=0.4]{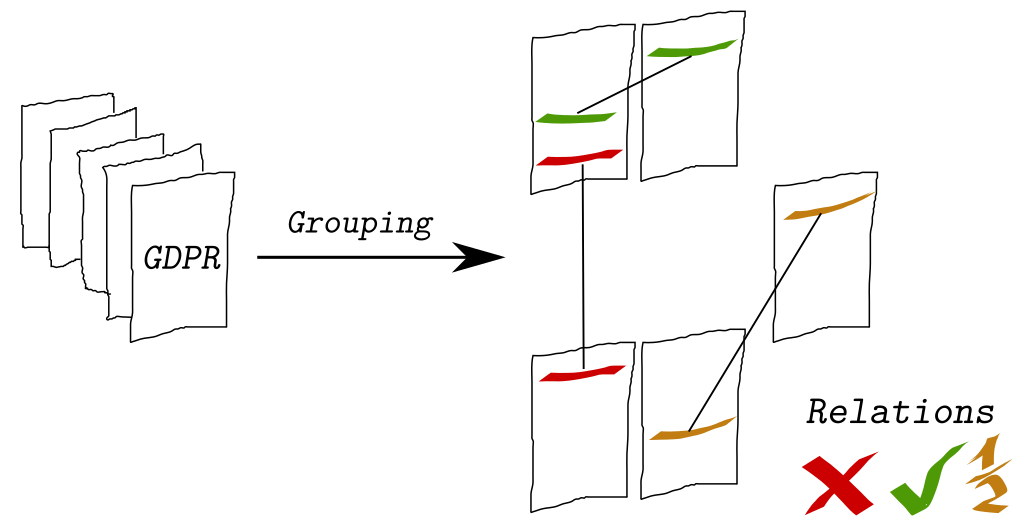}
    \caption{Extraction and Grouping of Constraints from Regulatory Documents\label{fig:overview}}
\end{figure}

We aim at facilitating the handling of complicated and extensive regulatory documents such as legal texts. Therefore, we have developed a method, that is able to structure the documents accordingly and to detect relations between sentences. As a result, similar sentences are highlighted which reduces the reading effort. In addition, a grouping based on, e.g., given topics serves to identify text passages that are relevant for a specific user. Figure \ref{fig:overview} outlines the basic idea. 
In addition, already implemented documents can also be integrated in order to detect conflicts and (partly) overlaps between sentences stemming from recently added documents. This supports the analysis of evolving regulatory documents over time.

The ConRelMiner method \cite{WiRi2018} consists of three steps, i.e., pre-processing, processing, and post-processing. 

The pre-processing step contains typical steps such as stemming and removal of stopwords. Novel is the fragmentation of the documents as described in \cite{DBLP:conf/otm/WinterRGFM17} where documents can be split along a certain semantics, e.g., paragraphs. As shown in \cite{DBLP:conf/otm/WinterRGFM17} this already enables a characterization of the documents, i.e.,  it can be derived which paragraph is associated with which theme or topic. Then the sentences that contain constraints are filtered out by using signal words. 

The processing step employs techniques from text mining \cite{aggarwal2012mining} and Natural Language Processing (NLP) \cite{nazir2017applications}, but also comprises novel concepts such as grouping sentences along topics and determining relations between the sentences based on their similarity. 
The grouping of sentences, resp. constraints can be customized individually depending on the type or size of the documents as well as additionally available information. In particular, a user can chose from three different methods. The first method uses term frequencies, whereas the second one exploits the structure of sentences and the third enables the integration of domain knowledge. Currently supported relations are ``redundant'', ``subsumed'', and ``conflicting''. For relations ``redundant'' and ``subsumed'' the related constraints can be viewed together and merged where applicable. Conflicting constraints can be also of interest, for example, if constraints contradict corresponding constraints in previous versions.

The result of the ConRelMiner method is a graph, which reflects the ordering by topics. In addition, sentences (the nodes of the graph) can be connected by edges describing the relations between them. Sentences can be ``redundant'' (marked green in the first figure, labeled with r in the graph), ``subsumed'' (orange, resp. s) or can be ``conflicting'' (red, resp. c).

\section{Application to GDPR}
\begin{figure}
    \centering
    \includegraphics[scale=0.4]{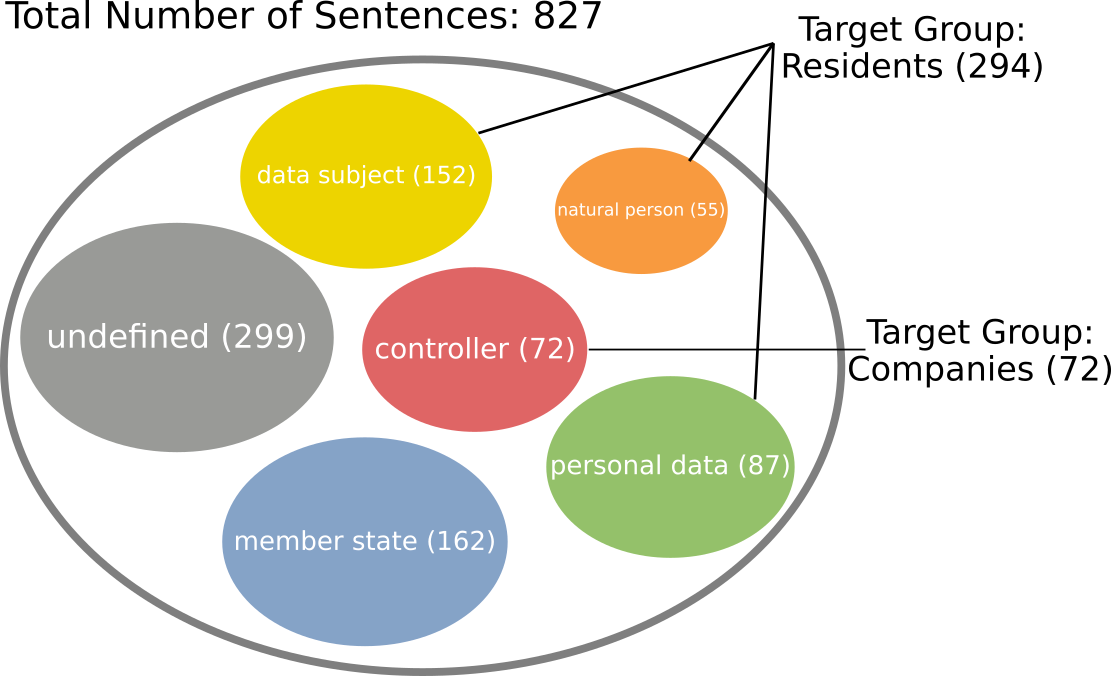}
     \caption{Grouping GDPR Constraints Along Target Groups\label{fig:grouping}}
\end{figure}

A currently very important regulation is the GDPR. This legislation consist of $88$ pages and is therefore quite extensive; however, citizens should know their rights regarding data privacy. Besides citizens, companies might also be interested in their duties regarding data privacy of customers. But not every paragraph is equally important for these distinct target groups. For example, the GDPR contains instructions how member states have to enact and adapt this law. These parts of the GDPR might be less relevant for companies or citizens. When applying the presented method, a filtering and grouping based on topics is possible which enables the direct detection of relevant passages for each target group. In this case, we have performed a grouping based on the words ``member state'' (162), ``natural person'' (55), ``data subject'' (152), ``personal data'' (87) and ``controller'' (72), inspired by the roles mentioned in \cite{DBLP:conf/sac/DeboisHLU17}. The number in brackets corresponds to the number of sentences we received. An overview of these results is illustrated by Fig. \ref{fig:grouping}. 

\begin{figure}
    \centering
    \includegraphics[scale=0.8]{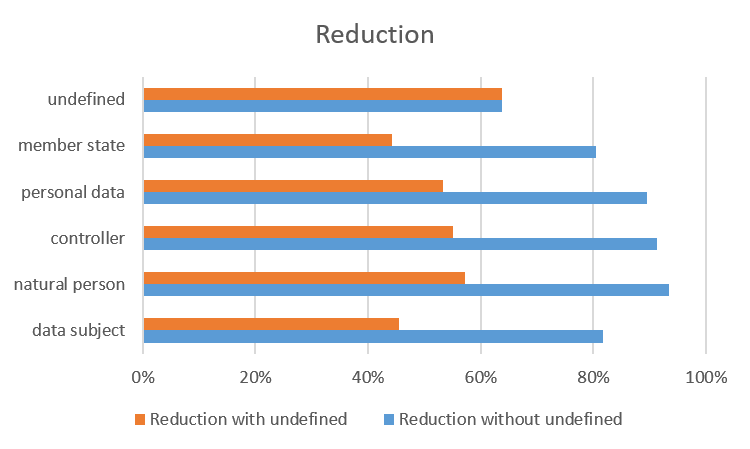}
    \caption{Reductions by Target Group}\label{fig:reductions}
\end{figure}

299 sentences did not contain any of the given words and were therefore categorized as ``undefined''. This sums up to 827 sentences. The ones that are relevant for citizens, i.e., those containing the words ``natural person'', ``data subject'' and ``personal data'' are altogether 294 sentences. Including the non-categorized sentences reduces the amount of sentences that need to be read from 827 to just 593, which corresponds to a reduction by approximately $28$\%. Assume that one is interested in what a company's controller needs to take care of. Without considering ``undefined'' sentences, only 72 sentences need to be evaluated, resulting in a reduction of $91$\%. If the ``undefined'' ones are considered  371 sentences have to be reviewed, still resulting in a reduction of $55$\%. The highest reduction of $93$\% can be achieved for target group ``natural person'' without ``undefined '' and $57$\% with ``undefined''. Note that  taking ``undefined'' into account for one target group is a maximum assumption in the sense that all ``undefined'' constraints actually belong to exactly this target group.   Figure \ref{fig:reductions} summarizes all reduction times.

 To explain this in more detail consider the following sentences from the GDPR:

\begin{itemize}
    \item \textsl{Where personal data are processed for the purposes of direct marketing, the data subject should have the right to object to such processing, including profiling to the extent that it is related to such direct marketing, whether with regard to initial or further processing, at any time and free of charge.}
    \item \textsl{Where personal data are processed for direct marketing purposes, the data subject shall have the right to object at any time to processing of personal data concerning him or her for such marketing, which includes profiling to the extent that it is related to such direct marketing.}
\end{itemize}

These are displayed as redundant sentences in the output graph and can therefore be handled at once. Reading the document in a chronological order, the first sentence would be on page 13, the second on page 45. It might be difficult to recognize that these are redundant sentences.

\begin{figure}
    \centering
    \includegraphics[scale=0.4]{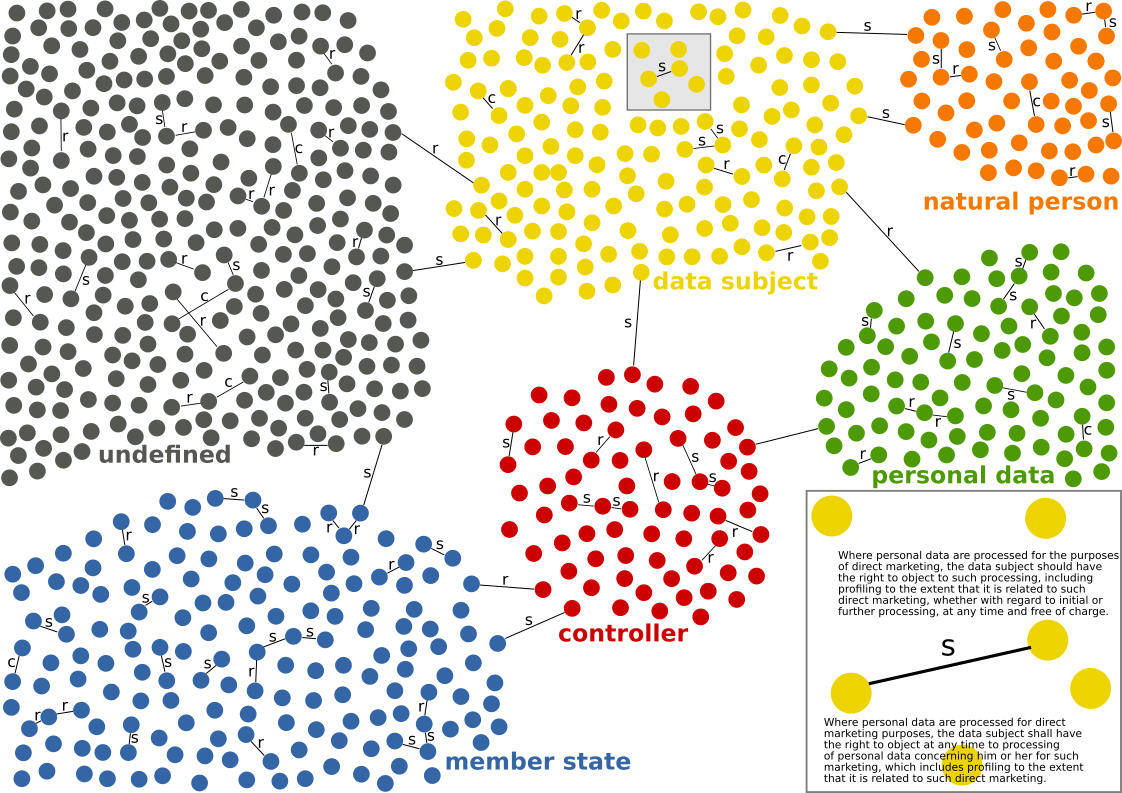}
    \caption{GDPR Analysis: Constraints and their Relations}\label{fig:graph}
\end{figure}

Grouping the GDPR as explained before results in the graph displayed schematically in Fig. \ref{fig:graph}.

Determining the relations between the constraints does not directly reduce the number of sentences that need to be read, but facilitates the implementation and maintenance of regulatory packages such as the GDPR. 

\section{Related Work}

Extracting knowledge from text is a broad and highly regarded task in science and practice. One distinction is the type of text or documents that is analyzed. It ranges from social media texts, e.g.,  \cite{DBLP:books/sp/social2011/AggarwalW11} over regulatory documents, e.g., \cite{bajwa2011sbvr,dragoni2016combining,WiRi2018,DBLP:conf/otm/WinterRGFM17} and business process descriptions, e.g., \cite{friedrich2011process,ghose2007process,riefer2016mining} to historic text analysis as in digital humanities, e.g., \cite{DBLP:conf/cikm/DonZGTACSP07}. 

Only few approaches target at digitalizing the GPDR such as \cite{DBLP:conf/sac/DeboisHLU17}: here the GDPR is formalized in terms of a declarative notion, the so called DCR graphs. 

\section{Future Challenges}

For future work we target to provide our tool as a web service. The input are the regulatory documents. During the application of the ConRelMiner the user can set different parameters for pre-processing and choose between different methods for the processing. This is particularly helpful if users already have some (domain) knowledge about the regulatory documents at hand (e.g., which additional information can be used). However, the ConRelMiner can be also applied without any prior knowledge and without any interaction: just input some documents and receive the filtered and grouped set of constraints, together with their relations. 

\section*{Acknowledgment} 
\vspace{-0.3cm}
This work has been funded by the Vienna Science and Technology Fund (WWTF) through project ICT15-072.

\bibliographystyle{abbrv}
\bibliography{references}

\end{document}